\documentclass[letterpaper]{article} % DO NOT CHANGE THIS
\usepackage{aaai2026}
\usepackage{times}  % DO NOT CHANGE THIS
\usepackage{helvet}  % DO NOT CHANGE THIS
\usepackage{courier}  % DO NOT CHANGE THIS
\usepackage[hyphens]{url}  % DO NOT CHANGE THIS
\usepackage{graphicx} % DO NOT CHANGE THIS
\usepackage{natbib}  % DO NOT CHANGE THIS AND DO NOT ADD ANY OPTIONS TO IT
\usepackage{caption} % DO NOT CHANGE THIS AND DO NOT ADD ANY OPTIONS TO IT
\usepackage{algorithm}
\usepackage{algorithmic}

\usepackage{amsmath}
\usepackage{amssymb}
\usepackage{mathtools}
\usepackage{amsthm}
\usepackage{multirow}
\usepackage{diagbox}
\usepackage{booktabs}

\newcommand{\bigO}[1]{\mathcal{O}\left(#1\right)}
\newcommand{\similarity}[1]{\text{sim}\left(#1\right)}
\newcommand{\e}[1]{\exp\left(#1\right)}
\newcommand{\fe}[1]{\phi\left(#1\right)}
\newcommand{\tmmathbf}[1]{\ensuremath{\boldsymbol{#1}}}

\usepackage{newfloat}
\usepackage{listings}
\DeclareCaptionStyle{ruled}{labelfont=normalfont,labelsep=colon,strut=off} % DO NOT CHANGE THIS
\floatstyle{ruled}
\newfloat{listing}{tb}{lst}{}
\floatname{listing}{Listing}
\title{SAOT: An Enhanced Locality-Aware Spectral Transformer for Solving PDEs} 

\makeatletter
\def\@fnsymbol#1{$\dagger$}
\makeatother

\author{
Chenhong Zhou\textsuperscript{\rm 1},
Jie Chen\textsuperscript{\rm 1}\thanks{Corresponding author.},  
Zaifeng Yang\textsuperscript{\rm 2} 
}
\affiliations {
    \textsuperscript{\rm 1}Department of Computer Science, Hong Kong Baptist University, Hong Kong SAR, China\\
    \textsuperscript{\rm 2}Institute of High Performance Computing (IHPC), Agency for Science, Technology and Research (A*STAR), Singapore\\
    20482795@life.hkbu.edu.hk,
    chenjie@comp.hkbu.edu.hk,
    yang\_zaifeng@a-star.edu.sg
}
\usepackage{bibentry}
\begin{document}

\maketitle

\begin{abstract}
Neural operators have shown great potential in solving a family of Partial Differential Equations (PDEs) by modeling the mappings between input and output functions. Fourier Neural Operator (FNO) implements global convolutions via parameterizing the integral operators in Fourier space. However, it often results in over-smoothing solutions and fails to capture local details and high-frequency components. To address these limitations, we investigate incorporating the spatial-frequency localization property of Wavelet transforms into the Transformer architecture. We propose a novel Wavelet Attention (WA) module with linear computational complexity to efficiently learn locality-aware features. Building upon WA, we further develop the Spectral Attention Operator Transformer (SAOT), a hybrid spectral Transformer framework that integrates WA’s localized focus with the global receptive field of Fourier-based Attention (FA) through a gated fusion block. Experimental results demonstrate that WA significantly mitigates the limitations of FA and outperforms existing Wavelet-based neural operators by a large margin. By integrating the locality-aware and global spectral representations, SAOT achieves state-of-the-art performance on six operator learning benchmarks and exhibits strong discretization-invariant ability.
\end{abstract}

\begin{links}
    \link{Code}{https://github.com/chenhong-zhou/SAOT}
    % \link{Datasets}{https://aaai.org/example/datasets}
    % \link{Extended version}{https://aaai.org/example/extended-version}
\end{links}

\section{Introduction}
\label{sec:introduction}

Solving partial differential equations (PDEs) is a fundamental and important task in real-world applications, enabling people to understand complex physical systems and make reliable decisions \cite{roubivcek2013nonlinear, azizzadenesheli2024neural}. Recently, deep learning has emerged as a promising tool for learning solution operators for PDEs, mainly attributed to its powerful modeling ability.  Versatile neural operators have been trained to learn the mappings between input and output function spaces, and then they can generalize well to inputs unseen during training~\cite{Li2020NeuralOG, lu2021learning}. This learning paradigm facilitates much faster inference than traditional numerical methods and shows great potential in surrogate modeling of PDEs  \cite{ li2021fourier,hao2024dpot,yue2024point}. Specifically, as a representative architecture of neural operators, Fourier neural operator (FNO) \cite{li2021fourier} parameterizes the integral kernel in Fourier space. This pioneering work has achieved impressive results in a wide range of applications \cite{ jiang2023efficient,liu2024difffno,li2024component,cho2024operator,leng2025fnin,liu2025aerogto} and sparked the surge of a series of follow-up methods like Geo-FNO \cite{Li2022FourierNO}, U-FNO \cite{Wen2021UFNOA}, F-FNO \cite{anonymous2023factorized}, AFNO \cite{guibas2021adaptive}, etc. Typically, FNO and its variants compute a forward Fast Fourier Transform (FFT), perform a matrix multiplication between Fourier modes and a learnable weight tensor, and conduct an inverse FFT, which amounts to global convolution in the spatial domain according to the convolution theorem \cite{li2021fourier, Li2022FourierNO, poli2022transform}. Hence, these methods can capture long-range dependencies by exploiting the strengths of learning in the frequency domain. However, such global convolution operations often lead to over-smoothing solutions and have deficiencies in capturing local and high-frequency details \cite{bachman2000fourier, tripura2022wavelet,oommen2024integrating}, which exacerbates the spectral bias phenomenon \cite{rahaman2019spectral,cao2021towards}. Some works have tried to replace Fourier transform with Wavelet transform to better capture local variations and nuances, such as MWT \cite{Gupta2021MultiwaveletbasedOL} and WNO \cite{tripura2022wavelet}. Wavelet basis functions, being localized in both space and frequency, are well-suited for capturing intricate local details and handling spatially discontinuous or abrupt signals, which are capabilities that the Fourier transform inherently lacks \cite{bachman2000fourier,khan2018learning}. Nevertheless, current implementations of Wavelet-based neural operators fail to achieve expected approximation accuracy, a shortcoming becoming significantly apparent against the rising dominance of Transformer-based models.

As a powerful architecture, Transformer \cite{NIPS2017_3f5ee243} has attracted growing interest and has been progressively adopted in operator learning \cite{guibas2021adaptive, kovachki2023neural, hao2024dpot}. The core of Transformers is the self-attention mechanism for effective token mixing to learn the relationships among tokens \cite{guibas2021adaptive,lee2021fnet}. However, the standard dot-product self-attention makes it computationally infeasible to model intricate correlations of massive mesh points, due to its quadratic complexity with respect to the sequence length \cite{hao2023gnot,wu2024Transolver}. Recently, Fourier transforms have been integrated into Transformer architectures to reduce computational complexity, such as FNet \cite{lee2021fnet}, GFNet \cite{rao2021global}, and AFNO \cite{guibas2021adaptive}. These different Fourier-based attentions have been proposed to replace standard self-attention and enable token mixing in the Fourier domain, but the aforementioned limitations still exist. To mitigate these issues, we investigate introducing the Wavelet transform into a Transformer architecture and propose a novel Wavelet-based attention mechanism.  To further combine the complementary benefits of the Fourier and Wavelet transforms, we propose a hybrid spectral Transformer framework that enhances the spectral representation for better approximation accuracy.

In this work, we propose a Wavelet Attention with linear complexity, aiming to capture local and high-frequency details while reducing computational expenses. Wavelet attention employs Fast Wavelet Transform (FWT) to decompose the input data into a sequence of wavelet subbands (coefficients), then applies a linearized attention mechanism between these wavelet subbands, and finally does inverse FWT. To fully harness the respective advantages of both Wavelet and Fourier transforms, we introduce a Spectral Attention Operator Transformer (SAOT) with the hybrid spectral attention. Specifically, a spectral attention layer consists of a Wavelet attention (WA) and a parallel Fourier attention (FA) as well as a gated fusion block to adaptively merge locality-aware representations from WA and global features from FA, thereby greatly boosting performance. Overall, our contributions are summarized as follows: 
\begin{itemize}
    \item  We propose WA with linear complexity to learn locality-aware representations by triggering self-attention learning between wavelet frequency subbands, which can compensate for the spectral deficiencies of FA. 
    \item We propose SAOT with Spectral Attention to adaptively adjust the contributions of global features from FA and local details from WA, which leverages the complementary advantages of two different frequency-domain operations.
    \item 
    Experimental results demonstrate that WA mitigates the loss of high-frequency details and outperforms current Wavelet-based neural operators by a large margin.
    By combining WA and FA, SAOT achieves state-of-the-art performance on several standard benchmarks and also presents an excellent discretization-invariant property. 
\end{itemize}

\section{Related Work}
\label{sec:related_work}
Neural operators aim to learn mappings between infinite-dimensional function spaces and therefore they are mesh-invariant, i.e., agnostic to the discretization of input and output functions \cite{lu2021learning,li2021fourier,kovachki2023neural}. With the exception of some earlier works, such as DeepONet \cite{lu2019deeponet} and GNO \cite{Li2020NeuralOG}, current prevailing neural operator approaches can be broadly categorized into the following two classes.

\paragraph{Spectral-based Neural Operators} 
As a seminal work, FNO \cite{li2021fourier} pioneered learning operator mappings in spectral space, inspiring numerous works based on typical spectral techniques. Applying the convolution theorem, FNO parameterizes the kernel integral operator by computing a matrix multiplication between Fourier modes and a learnable weight, which is equivalent to performing global convolution in the spatial domain \cite{li2021fourier,guibas2021adaptive}. Subsequently, U-NO \cite{rahman2022u} and U-FNO \cite{Wen2021UFNOA} plug FNO into U-Net \cite{ronneberger2015u} to improve the performance.  Geo-FNO \cite{Li2022FourierNO} extends the applicability of FNO from uniform grids to arbitrary geometries. F-FNO \cite{anonymous2023factorized} factorizes the Fourier representation to process each spatial dimension independently. SNO \cite{Fanaskov2022SpectralNO} aims to reduce aliasing errors in FNO using basic spectral methods. However, global convolution implemented in FNO and its variants is often vulnerable to over-smoothing and neglects local features \cite{ tripura2022wavelet, liu2024neural}. To implement local receptive fields, local neural operators \cite{liu2024neural} are developed to learn integral operators with locally supported kernels. In addition, some works have utilized Wavelet transforms to better capture locality by introducing multiscale wavelet bases in MWT \cite{Gupta2021MultiwaveletbasedOL} and wavelet integral layers in WNO \cite{tripura2022wavelet}.

\paragraph{Attention-based Neural Operators}
Another class of operator learning approaches is based on Transformer architectures, as the attention mechanism provides large modeling capability and flexibility \cite{hao2023gnot}. The quadratic complexity of attention is a critical challenge when handling large-scale irregular grids in PDEs. Much effort has been devoted to tackling the complexity issue by adopting smaller patches during patchify operation \cite{wu2023LSM,wang2024bridging,hagnberger2024vectorized}, introducing Fourier-based attentions \cite{hao2024dpot,guibas2021adaptive,rao2021global,lee2021fnet}, and designing efficient attentions \cite{Cao2021ChooseAT,li2023transformer, li2023scalable, hao2023gnot, chen2024positional}. Specifically, AFNO \cite{guibas2021adaptive} adapts FNO as an efficient token mixer with quasi-linear complexity, which is applied for operator learning in DPOT \cite{hao2024dpot}. Galerkin-type attention \cite{Cao2021ChooseAT} is developed to remove softmax normalization and achieve linear scaling. Afterwards, OFormer \cite{li2023transformer}, GNOT \cite{hao2023gnot}, and ONO \cite{anonymous2023improved} employ the linear-complexity variants of attention \cite{kitaev2020reformer,Katharopoulos2020TransformersAR,choromanski2021rethinking} for learning operators. Furthermore, IPOT \cite{lee2024inducing}, PiT \cite{chen2024positional}, and Transolver \cite{wu2024Transolver} design novel attention mechanisms to improve model performance with affordable computational complexity.

\section{Method}
\label{sec:method}
To compensate for the deficiencies in Fourier-based attention, we first propose an efficient Wavelet attention that learns locality-aware features to enhance high-frequency details while reducing computational complexity. Next, we introduce a spectral attention layer to merge the features from a WA and a parallel FA via a gated fusion block. Finally, we present the overall architecture of SAOT and analyze its computational complexity.

\paragraph{Problem Setup} Let $\Omega\subset \mathbb{R}^{d}$ be a bounded open set. The input and output functions $a$ and $u$ are drawn from Banach spaces $\mathcal{A}$ and $\mathcal{U}$, respectively: $a\in\mathcal{A}(\Omega;\mathbb{R}^{d_a}), u\in\mathcal{U}(\Omega;\mathbb{R}^{d_u})$. Suppose $\Phi$ is an operator mapping from input to output function, i.e., $\Phi:\mathcal{A} \to \mathcal{U}, \; a\mapsto u$. In practice, representing the function directly is hard. Hence, training data are the values of the input and output functions sampled on regular or irregular meshes $g$. Our goal is to learn a neural operator $\Phi_{\theta}$ to approximate $\Phi$ from a set of observed input-output pairs, where $\theta$ denotes the trainable parameters.

\paragraph{Preliminary: Self-attention}
Given the input sequence $X= \{\tmmathbf{x}_{j}\}_{j=1}^{N}\in\mathbb{R}^{N\times D}$ which contains $N$ elements and each element has $D$ feature channels,  self-attention employs three trainable matrices $W^q, W^k, W^v\in \mathbb{R}^{D \times D}$ to obtain the corresponding representations ($Q$: queries, $K$: keys, and $V$: values): 
\begin{equation}
\label{eq:attention-weights}
\quad \tmmathbf{q}_{j} = \tmmathbf{x}_{j} W^q, 
\quad \tmmathbf{k}_{j} = \tmmathbf{x}_{j} W^k,\quad \tmmathbf{v}_{j} = \tmmathbf{x}_{j} W^v,
\end{equation} 
where $\tmmathbf{q}_{j}$, $\tmmathbf{k}_{j}$, and $\tmmathbf{v}_{j}$ are the $j$-th rows of $Q$, $K$, and $V$, respectively.  
Softmax attention is a specific form of self-attention, where the similarity score is computed as the exponential of a dot-product between queries and keys, which can be formulated as follows: 
\begin{equation}
	\label{eq:selfatt}
\tmmathbf{s}_i =  \frac{\sum_{j=1}^N \exp (\tmmathbf{q}_i \cdot \tmmathbf{k}_j /\tau) \tmmathbf{v}_j}{\sum_{l=1}^N \exp (\tmmathbf{q}_i \cdot \tmmathbf{k}_l / \tau)}, 
\end{equation}
where $\tau$  is a temperature hyperparameter and $\tmmathbf{s}_i$ denotes the $i$-th row  vector of attention matrix $S\in\mathbb{R}^{N\times D}$. The computational cost of self-attention is scaling quadratically w.r.t. the sequence length, i.e., $\bigO{N^2}$. A generalized self-attention function for any similarity function \cite{katharopoulos2020transformers} can be expressed as: 
\begin{equation}
   \tmmathbf{s}'_i =  \frac{\sum_{j=1}^N \similarity{\tmmathbf{q}_i,  \tmmathbf{k}_j} \tmmathbf{v}_j}{\sum_{l=1}^N \similarity {\tmmathbf{q}_i,  \tmmathbf{k}_l}} ,
    \label{eq:general-attn}
\end{equation}
If the similarity function is $\similarity{\tmmathbf{q}, \tmmathbf{k}} = \e{\tmmathbf{q} \cdot \tmmathbf{k}/\tau}$, the generalized form in Eq. \ref{eq:general-attn} reduces to the softmax attention in Eq. \ref{eq:selfatt}. Previous works have demonstrated that a kernel with non-negative similarity scores can be used to replace the exponential kernel in softmax attention and linearize the attention to reduce the computational complexity \cite{ Cao2021ChooseAT,choromanski2021rethinking,katharopoulos2020transformers}.

\begin{figure*}
    \centering
    \includegraphics[width=0.95\linewidth]{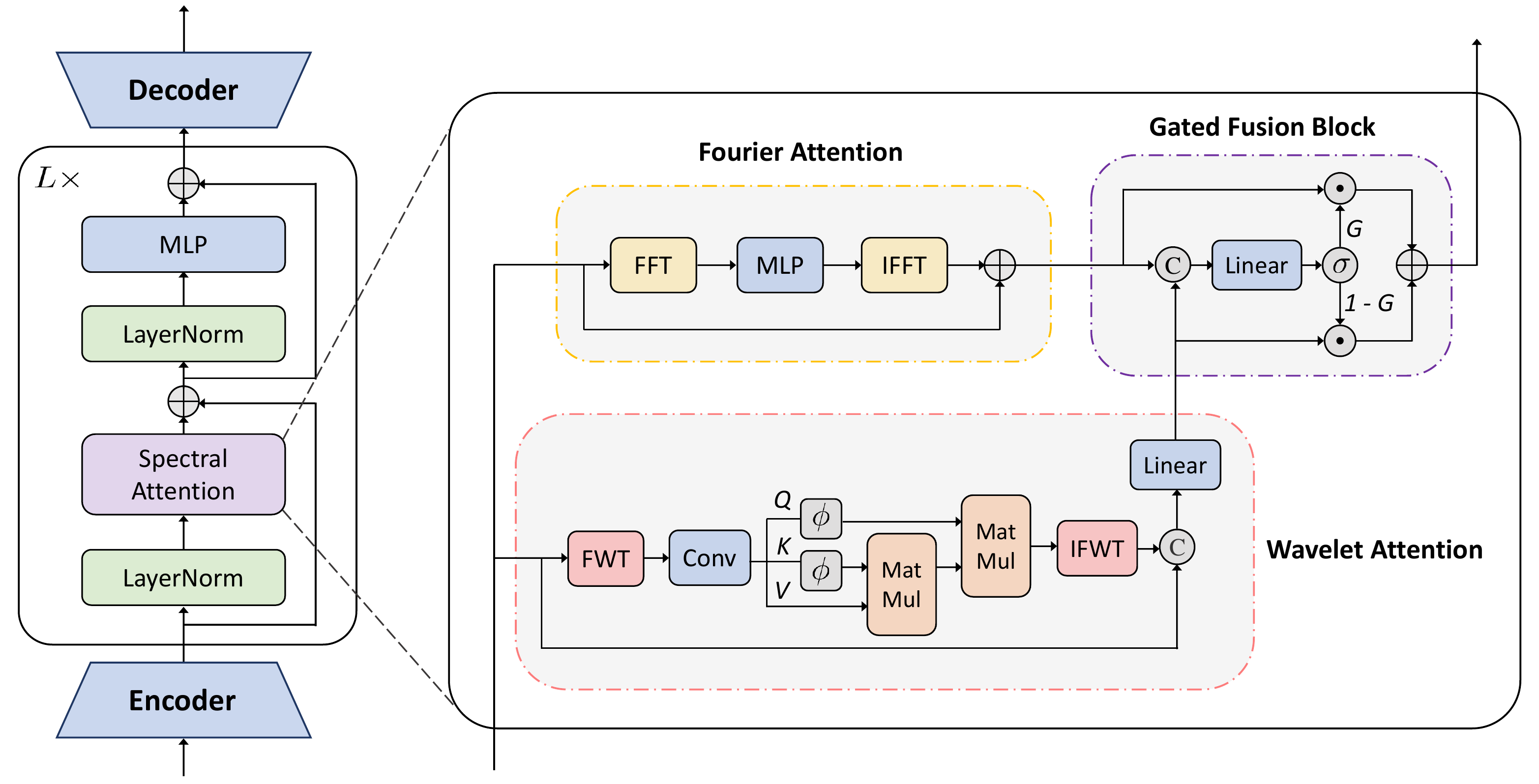}
    \caption{Architecture of the SAOT model. It contains three main components: an encoder, a processor with $L$ pre-norm Transformer blocks, and a decoder. The core of our Transformer block is the proposed spectral attention, which is composed of a Fourier Attention (FA) and a Wavelet Attention (WA), as well as a gated fusion block.
      }
    \label{fig_model}
\end{figure*}
 
\subsection{Wavelet Attention (WA)} 
An important benefit of the Wavelet transform is providing spatial-frequency localized representations; thus, we propose the Wavelet Attention, an efficient implementation for mixing tokens in the Wavelet domain. Formally, given a mesh set with $N$ points , we consider  $\Omega\subset \mathbb{R}^{2}$ as a depicted example and thus $N = H\times W$. The inputs are projected into deep feature maps  $X\in \mathbb{R}^{H \times W \times D}$. We first embed $X$ into $\bar{X}\in \mathbb{R}^{H \times W \times{\frac{D}{4}}}$  via a convolution layer to reduce channel dimension before employing Fast Wavelet Transform (FWT). Then we use FWT to decompose $\bar{X}$ into different wavelet subbands (coefficients) with common smaller sizes and different frequency components. Here Haar wavelet is chosen for FWT as in \cite{liu2020wavelet,yao2022wave} for simplicity. Concretely, the high-pass filter $f_H = (1/\sqrt 2, -1/\sqrt 2)$ and low-pass filter $f_L = (1/\sqrt 2, 1/\sqrt 2)$ in FWT are applied along rows to transform $\bar{X}$ into subbands $X_H$ and $X_L$.  Then, these filters are applied to the columns of $X_H$ and $X_L$, resulting in four distinct wavelet subbands: $X_{HH}$, $X_{HL}$, $X_{LH}$, and $X_{LL}$, where each subband has dimensions of  ${{\mathbb{R}}^{{\frac{H}{2}} \times {\frac{W}{2}} \times {\frac{D}{4}}}}$. $X_{LL}$ denotes the low‐frequency component in both dimensions, capturing fundamental features at a coarse level. $X_{LH}$, $X_{HL}$, and $X_{HH}$ signify the high‐frequency components, preserving the details at a fine level. These wavelet subbands collectively form a compact and complete representation of the original data at different levels of detail and orientation. Hence, we concatenate them along the channel dimension to form  $\widetilde{X}=\operatorname{Concat}(X_{LL}, X_{LH}, X_{HL}, X_{HH}) \in {{\mathbb{R}}^{{\frac{H}{2}} \times {\frac{W}{2}} \times D}}$. Notice that the spatial dimensions are halved in both height and width, resulting in a one-quarter reduction in total, which lowers computational demands for the subsequent attention calculation.

By leveraging a kernel feature map $\fe{x}$ to linearize the attention mechanism, we can rewrite Eq. \ref{eq:general-attn} and further simplify it via the associative property of matrix product: 
\begin{align}
 \begin{small}
 \label{eq:asso-attn}
  \tmmathbf{s}'_i   = \frac{\sum_{j=1}^N \fe{\tmmathbf{q}_i}^T \fe{\tmmathbf{k}_j} \tmmathbf{v}_j}{\sum_{l=1}^N \fe{\tmmathbf{q}_i}^T \fe{\tmmathbf{k}_l}} = \frac{\fe{\tmmathbf{q}_i}^T \left (\sum_{j=1}^N \fe{\tmmathbf{k}_j} \otimes \tmmathbf{v}_j \right ) } {\fe{\tmmathbf{q}_i}^T \sum_{l=l}^N \fe{\tmmathbf{k}_l}}.
  \end{small}
\end{align}
According to Eq. \ref{eq:asso-attn},  $\sum_{j=1}^N\fe{\tmmathbf{k}_j}\otimes  \tmmathbf{v}_j$ and $\sum_{l=1}^N \fe{\tmmathbf{k}_l}$ can be computed once and efficiently reused across all queries, so that such an attention has efficient computational and memory complexity $\bigO{N}$.  Here we use the feature map as in \cite{katharopoulos2020transformers}: $\fe{x} = \text{elu}(x) + 1$, where $\text{elu}(\cdot)$ denotes the exponential linear unit activation function \cite{clevert2015fast}. Finally, we use $\operatorname{LinearAttn}$ to represent this linearized attention. Before performing self-attention learning over wavelet subbands, we optionally apply a convolution operation like a kernel of 3$\times$3 to $\widetilde{X}$ to enforce spatial locality, thereby generating the locally contextualized features $X^w$ \cite{yao2022wave}. Next, $X^w$ is reshaped and linearly transformed into queries/keys/values: $Q^w, K^w, V^w \in \mathbb{R}^{n \times D}$, where $n = {\frac{H}{2}} \times {\frac{W}{2}}$. Then we employ the linearized attention mechanism to capture the long-range contextualized information among wavelet subbands: $S' = \operatorname{LinearAttn}(Q^w, K^w, V^w)$.

Afterwards, we reshape $S'$ into $\mathbb{R}^{\frac{H}{2} \times \frac{W}{2} \times D}$ and apply the inverse FWT (IFWT) to obtain the reconstruction $X^r \in \mathbb{R}^{H \times W \times \frac{D}{4}}$. Finally, the reconstructed and original data are concatenated, followed by a linear embedding layer to compose the final output of WA: i.e.~$X^{\text{WA}}=\operatorname{Linear}\left(\operatorname{Concat}(X, X^r)\right)$. As a result, the main operations in the WA can be summarized as $\operatorname{FWT-Conv-LinearAttn-IFWT}$. This attention mechanism harnesses the benefits of wavelets to accomplish local receptive fields and significantly reduce the complexity.

\subsection{Spectral Attention Layer} 
We propose a spectral attention layer that integrates a WA and a FA in parallel, combined with a gated fusion block to synergize the advantages of Wavelet and Fourier transforms.

\paragraph{Fourier Attention (FA)} Here, we utilize the FA similar to AFNO \cite{guibas2021adaptive}, which adapts FNO for token mixing in Transformer architectures. 
Unlike AFNO, our FA omits sparsity operations to retain all frequency modes, thereby preserving the full expressivity of the representation \cite{hao2024dpot}. Concretely, FA performs the global convolution by approximating the kernel integral operator in Fourier space \cite{li2021fourier,guibas2021adaptive}: 
\begin{equation}
  \mathcal{K} (X) (g) =\mathcal{F}^{- 1} (R_{\psi} \cdot
  \mathcal{F} (X)) (g) ,\qquad \forall g \in \Omega
  \label{FA_eqn}
\end{equation}
where $\mathcal{K}$ is a kernel integral operator parameterized by $R_{\psi}$, $\mathcal{F}$, $\mathcal{F}^{- 1}$ denote the Fourier transform and its inverse, and $X$ denotes the input function discretized on mesh $g$.  Specifically, $R_{\psi}$ represents block-wise MLP layers instead of a learnable complex-valued weight tensor in FNO to dramatically reduce training parameters and memory \cite{guibas2021adaptive}.  Thus, the FA can be summarized as $\operatorname{FFT-MLP-IFFT}$, and its output is denoted as $X'$. Following  AFNO, we add a residual term to $X'$ to generate the final output of FA:  $X^{\text{FA}} = X + X'$. The FA module is running in parallel to the proposed WA.

\paragraph{Gated Fusion Block}  
To compensate for local details, we incorporate a WA to complement FA and introduce a gated fusion block to adaptively merge the features from two spectral operations. Concretely, given the features $X^{\text{FA}} \in \mathbb{R}^{N \times D}$ and $X^{\text{WA}} \in \mathbb{R}^{N \times D}$ derived from FA and WA respectively, we first concatenate them along the channel dimension and then feed them into a linear embedding layer, followed by a sigmoid activation function:
\begin{equation} 
G = \sigma\left( \operatorname{Linear}\left( 
\operatorname{Concat}(X^{\text{FA}}, X^{\text{WA}})
\right) \right),
\label{eq:gating_map}
\end{equation}
where \( \sigma(\cdot) \) is the sigmoid activation function, and  $G \in \mathbb{R}^{N \times D}$ denotes the generated gating map. Next, this gating map can be used to fuse the features by performing an element-wise weighted sum: 
\begin{equation}
X^{\text{SA}} = G \odot X^{\text{FA}} + \left( 1 - G \right) \odot  X^{\text{WA}}, %\otimes
\label{eq:fused_features}
\end{equation}     
where \( \odot \) denotes element-wise multiplication, and $X^{\text{SA}}$ denotes the fused features, i.e., the final output of a spectral attention (SA) layer. As a result, the SA layer integrates global dependencies from a FA and local features from a WA to leverage the complementary advantages of these two frequency-domain operations.

\begin{table*}[t]
	\centering
			\renewcommand{\multirowsetup}{\centering}
			\setlength{\tabcolsep}{7pt}
			\begin{tabular}{lccccccc}
				\toprule
			     Benchmarks    & Input & Output  & Geometry & $N$ & $N_t$  & Train set  & Test set    \\ 
                  \midrule
                   Darcy     & Diffusion coefficient  & Fluid pressure &Regular grid & $85 \times 85$ &-  &1000 &200   \\ 
		         NS  & Past velocity & Future velocity  &Regular grid  & $64 \times 64$ &10   &1000 &200  \\	     
                % \midrule
			     Airfoil  & Mesh points  & Mach number   & Structured mesh & $221 \times 51$ &-   &1000 &200 \\ 
        	  Pipe  & Mesh points & Fluid velocity   & Structured mesh & $129 \times 129$ &-   &1000 &200  \\ 
               Plasticity  & Mesh points & Mesh deformation   & Structured mesh  & $101 \times 31$ &20  &900 &80  \\  
             Elasticity  & Structure & Inner stress  & Point cloud & 972 &-  &1000 &200 \\
				\bottomrule
			\end{tabular}
  	\caption{Summary of benchmarks. $N$ means the spatial resolution, and $N_t$ means temporal dimension. The last two columns denote the number of samples in the train and test sets.
    } 
    \label{tab_dataset_detail}
\end{table*}

\subsection{Spectral Attention Operator Transformer (SAOT)} 
In general, our SAOT model is built upon a Transformer architecture where the standard softmax attention is replaced with the proposed spectral attention layer.

\paragraph{Overall Architecture}
Specifically, SAOT adopts the typical Encoder-Processor-Decoder architecture:
\begin{equation}
\Phi_{\theta}=\text{Decoder}\circ \text{LAYER}_L\circ\cdots\circ \text{LAYER}_1\circ \text{Encoder},
\end{equation}
where the encoder is usually implemented by linear layers to lift the input function from $\mathbb{R}^{d_a}$ to a high-dimensional feature space $\mathbb{R}^{D}$. As shown in Figure \ref{fig_model}, the processor is composed of $L$ stacked pre-norm Transformer blocks \cite{NIPS2017_3f5ee243, xiong2020layer}, and the $l$-th block can be formalized as follows:
\begin{equation}
\begin{split}\label{equ:overall}
\hat{X}^{l} &= \operatorname{SA}\left(\operatorname{LN}\left({X}^{l-1}\right)\right) + {X}^{l-1} \\
{X}^{l} &= \operatorname{MLP}\left(\operatorname{LN}\left(\hat{X}^{l}\right)\right) + \hat{X}^{l},
	\end{split}
\end{equation}
where $l=1,\cdots, L$, $\operatorname{LN}$ denotes layer normalization \cite{ba2016layer}, and $\operatorname{MLP}$ is usually implemented with a two-layer feedforward network. $X^{l}\in\mathbb{R}^{N\times D}$ is the output of the $l$-th block. $X^{0}\in\mathbb{R}^{N\times D}$ denotes the input deep features from the encoder. Finally, $X^{L}$ is further fed to the decoder to perform a linear projection to the output dimension $\mathbb{R}^{d_u}$.

\paragraph{Computational Complexity} 
Since a SA is composed of a FA and a WA running in parallel, its overall complexity is dominated by the higher-complexity component. Assuming $N \gg D$, the FA has the complexity of $\bigO{ND \log N}$ due to FFT. The core of WA is a linear-complexity variant of attention, and its complexity scales linearly with the sequence length as $\bigO{ND^2 }$. Hence, the complexity of a SA is  $\bigO{ \max\left(ND^2, ND\log N\right)}$. As a result, the total complexity for the SAOT with $L$ Transformer blocks is $\bigO{L\max\left(ND^2, ND\log N\right)}$.

\section{Experiments}
\label{sec:experiments}
In this section, we first introduce the experimental setup. Next, we compare the performance between SAOT and more than ten baselines on six standard benchmarks to demonstrate the effectiveness of SAOT. Finally, we conduct the ablation study, make more comparisons, and evaluate the model's generalization ability.

\subsection{Experimental Setup } 
\paragraph{Benchmarks} 
Our experiments include six standard operator learning benchmarks: Darcy, Navier-Stokes (NS), Airfoil, Pipe, Plasticity, and Elasticity. The first two benchmarks are from FNO \cite{li2021fourier}, and the others are from geo-FNO \cite{Li2022FourierNO}.
These problems encompass various tasks in fluid and solid physical systems governed by different PDEs. Data for these problems is collected on uniform regular grids or irregular grid structures. Here we give a summary of these benchmarks and follow the experimental settings in the previous works \cite{li2021fourier, Li2022FourierNO, wu2023LSM, wu2024Transolver}, which are summarized in Table \ref{tab_dataset_detail}. More detailed descriptions of these benchmarks can be found in the supplementary material.

\paragraph{Baselines}
We compare SAOT with various representative and powerful architectures as baselines: including \textbf{FNO} \cite{li2021fourier} and its three variants: \textbf{U-NO} \cite{rahman2022u}, \textbf{Geo-FNO} \cite{Li2022FourierNO}, and \textbf{F-FNO} \cite{anonymous2023factorized}; two Wavelet-based neural operators: \textbf{MWT} \cite{Gupta2021MultiwaveletbasedOL} and \textbf{WNO} \cite{tripura2022wavelet}; and five advanced attention-based operators:  \textbf{Galerkin} \cite{Cao2021ChooseAT}, \textbf{OFormer} \cite{li2023transformer}, \textbf{GNOT} \cite{hao2023gnot}, \textbf{IPOT} \cite{lee2024inducing}, and \textbf{Transolver} \cite{wu2024Transolver}. 
Concretely, FNO and U-NO are only applicable to the Darcy and Navier-Stokes datasets since they are limited to regular grids. Geo-FNO extends the application of FNO to irregular grid datasets. MWT and WNO are two typical neural operators based on the Wavelet transform. The last five Transformer-based models design different attention mechanisms to handle complex geometries and enhance the performance. It’s noteworthy that Transolver is the latest state-of-the-art model for operator learning.

\begin{table*}[t]
	\centering
	\setlength{\tabcolsep}{16pt}
			\begin{tabular}{ccccccc} 
       \toprule
       \bf{Model} & Darcy & NS   & Airfoil  & Pipe  & Plasticity  & Elasticity \\
      \midrule
        FNO \shortcite{li2021fourier} & 0.0108          & 0.1556    &/  &/ &/ &/ 
        \\
        U-NO* \shortcite{rahman2022u}  & 0.0091     & 0.1408    &/  &/ &/ &/  \\
        Geo-FNO \shortcite{Li2022FourierNO}  & 0.0108      & 0.1556     & 0.0138 & 0.0067 & 0.0074    &0.0229 \\
         F-FNO \shortcite{anonymous2023factorized}  & 0.0077    & 0.2322   & 0.0078  & 0.0070 & 0.0047   & 0.0263 \\
         MWT* \shortcite{Gupta2021MultiwaveletbasedOL} & 0.0067   & 0.1553   & 0.0076   & 0.0072  & 0.0027  & 0.0334    \\
        WNO* \shortcite{tripura2022wavelet}   & 0.0242   & 0.1613   & 0.0188   & 0.0070  & /  & 0.0465   \\
       \midrule
        Galerkin \shortcite{Cao2021ChooseAT}    & 0.0084         & 0.1401    & 0.0118  & 0.0098 & 0.0120 & 0.0240  
        \\
        Oformer \shortcite{li2023transformer}    & 0.0124        & 0.1705    & 0.0183  & 0.0168 & 0.0017 & 0.0183 \\
        GNOT \shortcite{hao2023gnot} & 0.0105   & 0.1380   & 0.0076  & / & /  & 0.0086 \\
       IPOT \shortcite{lee2024inducing} & 0.0085    & \underline{0.0885}     & 0.0088  & /  & 0.0033  & 0.0156  \\
        Transolver* \shortcite{wu2024Transolver} & \underline{0.0058} & 0.0985  & \underline{0.0053}  & \bf{0.0043} & \underline{0.0012} &\bf{0.0067}      \\
    \midrule
      SAOT & \bf{0.0049}   & \bf{0.0688}     &\bf{0.0048}  &\underline{0.0063} & \bf{0.0008} & \underline{0.0080}   \\
      \bottomrule
        \end{tabular}
    \caption{Performance comparisons with several baselines across six standard benchmarks. A smaller value indicates better performance. The best result is \textbf{bolded}, and the second best result is \underline{underlined}.  ``*" means that the reported results are reproduced by us. ``/" means that the result for this benchmark is not available.  } 
    \label{table_results}
\end{table*}

\paragraph{Implementation}
For fair comparisons, all the experiments are conducted on one NVIDIA Tesla V100S-PCIE-32GB. 
Following the convention of related literature \cite{li2021fourier, Li2022FourierNO,wu2024Transolver}, our model is trained with relative $L^2$ loss for 500 epochs with an initial learning rate of $1 \times 10^{-3}$. The relative $L^2$ error is also used as the evaluation metric to measure the quality of predictions.

\subsection{Main Results}
Table \ref{table_results} reports the relative $L^2$ errors of our proposed SAOT and several baselines for six benchmarks. Except those of F-FNO, Galerkin, and Oformer cited from \cite{wu2024Transolver}, the results of other baselines are cited from those original papers or reproduced by ourselves marked as ``*".  If the baseline cannot apply to a benchmark or the original papers did not report the results for this benchmark, we mark it as ``$/$".

\begin{figure*}[t]
    \centering
    \includegraphics[width=\linewidth]{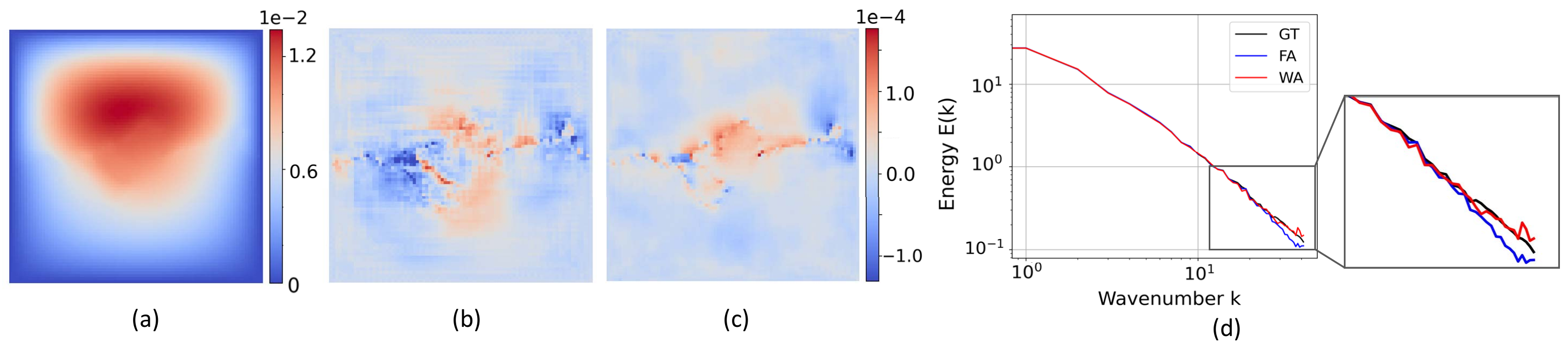}  %spectrum_fangda_no_caption.pdf
    \caption{(a) Ground truth (GT); (b) Prediction errors of Fourier Attention (FA); (c) Prediction errors of Wavelet Attention (WA); (d) The energy spectrum of GT and the predictions of FA and WA with respect to the wavenumber. WA yields better-aligned energy spectrum predictions with true distributions than FA, especially at higher wavenumbers.
    } 
    \label{fig_spectrum}
\end{figure*}

In Darcy and Navier-Stokes benchmarks, where the domains are discretized into uniform regular grids,  SAOT achieves the lowest prediction errors, outperforming all the baselines. Specifically, SAOT significantly surpasses the spectral-based neural operators listed in Table \ref{table_results} and also shows noticeable advantages over Transformer-based operators. In particular, compared to Transolver, SAOT gains remarkable improvement with relative promotions of 15.5\% (0.0058$\to$0.0049) in Darcy and 22.3\% (0.0985$\to$0.0688) in Navier-Stokes. In addition, SAOT exhibits excellent performance in Airfoil, Pipe, and Plasticity benchmarks. Especially, SAOT achieves remarkably lower errors than Transolver with relative performance promotions of 9.4\% and 33.3\% in Airfoil and Plasticity, respectively. The Elasticity benchmark, where the geometry is modeled using irregular point clouds, poses large challenges for operator learning. We can see that the above five spectral-based neural operators cannot or struggle to estimate the stress on unstructured point clouds, with errors exceeding 2\%. In contrast, the latest models (GNOT/IPOT/Transolver) and our SAOT show the superiority of Transformer-based backbone models in handling complex geometries. Specifically, SAOT achieves the competitive performance with a relative $L^2$ error of 0.8\% on this challenging benchmark. 
In addition to these quantitative comparisons, we also visualize the prediction results for an intuitive comparison in the supplementary material. Overall, SAOT exhibits its universal approximation capability across diverse datasets with complex geometries.

\subsection{Ablation Study} 
To validate the capability of WA in capturing high-frequency details, we perform the experiments on the Darcy dataset and compare the performance of models using FA alone versus WA alone in Figure \ref{fig_spectrum}. We can see that the prediction errors of WA are relatively closer to 0 than those of FA. More importantly, Figure \ref{fig_spectrum}(d) shows that FA generates the predictions whose energy aligns well with the ground truth for only lower wavenumbers, but it ignores the higher frequencies, which can explain why FA often tends to generate over-smoothing solutions. Fortunately, our proposed WA mitigates this spectral bias by offering predictions with a better alignment on high-wavenumber modes.

To verify the effectiveness of mixing tokens in different frequency domains, we employ different attentions in SAOT and report the results in Table \ref{table_ablation}. We can see that WA achieves better performance than FA, especially for Elasticity with a remarkably lower prediction error from 0.0232$\to$0.0129. As the subtle changes in a material structure can heavily impact the variations in stress distribution, the strength of WA to capture local details is beneficial for precise stress estimations and thus brings significant promotions. 
Furthermore, we find that WA shows a noticeably better performance than the two Wavelet-based neural operators: MWT and WNO in Table \ref{table_results}, verifying the superiority of the proposed Wavelet-based attention. Ultimately, SA further significantly reduces the prediction errors on both benchmarks by combining the complementary features from FA and WA.

\begin{table}
	\centering
			\setlength{\tabcolsep}{6pt}
           \renewcommand{\arraystretch}{1.15}
			\begin{tabular}{c|cc|cc}
          \toprule  
			\multirow{2}{*}{Attn.} & \multicolumn{2}{c|}{Darcy}   & \multicolumn{2}{c}{Elasticity} \\ 
            \cline{2-5}
                  &{Param. (M)} & {Rel. L2} & {Param. (M)} &{Rel. L2} \\
             \midrule
		       FA & 0.651  & 0.0058 & 0.576 & 0.0232 \\
                 WA & 2.361 & 0.0057  & 1.514 & 0.0129 \\
                  SA & 2.694 & \bf{0.0049} & 2.040 & \bf{0.0080} \\
         \bottomrule
			\end{tabular}
    \caption{Ablation study on the Darcy and Elasticity benchmarks about different attention mechanisms.} 
  \label{table_ablation}
\end{table}

\subsection{Comparison with Other Spectral Transformers} 
We further compare the proposed SAOT with three classical spectral Transformers: FNet \cite{lee2021fnet}, GFNet \cite{rao2021global}, and AFNO \cite{guibas2021adaptive}, which are proposed for vision tasks to perform token mixing in the Fourier domain. The results are shown in Table \ref{table3}. FNet employs an unparameterized Fourier Transform to replace self-attention, leading to the smallest parameter size, but it results in the largest errors on two benchmarks. Owing to an element-wise multiplication between Fourier features and learnable global filters, GFNet experiences a dramatic rise in trainable network parameters, especially for large-scale grids in the Darcy dataset, yet its performance does not scale significantly with the increased parameters. AFNO performs better than FNet given similar parameter counts, but its error on the Elasticity dataset is still greater than 2\%.
Our SAOT model combines the advantages of two types of frequency-domain attentions under a modest parameter count, and thus achieves the lowest errors on both datasets.

\begin{table}[t] 
	
	\centering
	
		\setlength{\tabcolsep}{6pt}
			 \renewcommand{\arraystretch}{1.15}
			\begin{tabular}{c|cc|cc}
            \toprule
			\multirow{2}{*}{Model} & \multicolumn{2}{c|}{Darcy}   & \multicolumn{2}{c}{Elasticity} \\
            \cline{2-5}
                  &{Param. (M)} &{Rel. L2} & {Param. (M)} &{Rel. L2} \\  
             \midrule
		       FNet  & 0.582  & 0.0168 & 0.466 &0.1044  \\
                 GFNet  & 8.067 & 0.0074  &1.869  &0.0230 \\
                 AFNO  & 0.651 & 0.0058  &0.576  &0.0228  \\ 
                 \midrule  
                  SAOT & 2.694 & \bf{0.0049} & 2.040 & \bf{0.0080} \\
		
        \bottomrule
			\end{tabular}
     \caption{Performance comparisons between SAOT and other popular spectral Transformers in terms of the parameter count and relative L2 error. 
 } 
    \label{table3}
\end{table}

\subsection{Generalization Ability under  Different Resolutions} 
\begin{figure}
\centering
\includegraphics[width=\linewidth]{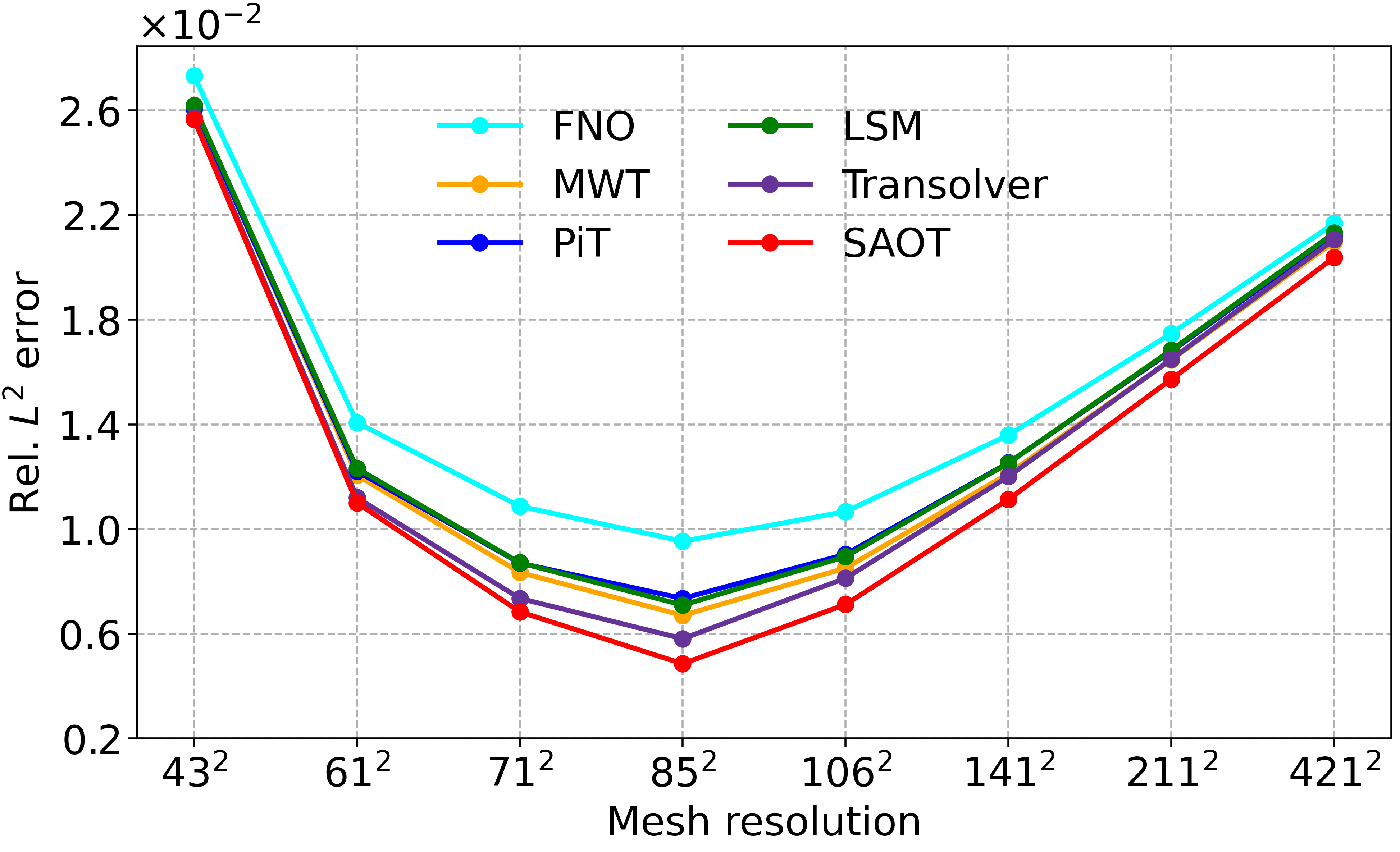}
    \caption{Model generalization performance on the Darcy dataset with different mesh resolutions.}
    \label{fig2}
\end{figure}

To evaluate the model's generalization ability, we conduct the zero-shot super-resolution evaluation on the Darcy benchmark. Specifically, we train the model with the data on a resolution of $85^2$ and then test the well-trained model with varying resolutions ranging from $43^2$ to $421^2$. Hence, we can assess the discretization-invariant property of our model and make comparisons with other methods: FNO \cite{li2021fourier}, MWT \cite{Gupta2021MultiwaveletbasedOL}, LSM \cite{wu2023LSM}, PiT \cite{chen2024positional}, and Transolver \cite{wu2024Transolver}. The comparison results are displayed in Figure \ref{fig2}. 
We can see that all the methods obtain the lowest errors when the test resolution matches the training resolution, i.e., $85^2$. This is reasonable because the model is trained to be good at extracting features from data of this resolution. Inconsistency in the resolution between training and test data can introduce additional variability and complexity, making it harder for the model to make accurate predictions. Therefore, as the test resolution moves further away from $85^2$, the test errors of all methods progressively rise, leading to U-shaped convex curves in Figure \ref{fig2}. Notably, our proposed SAOT model consistently yields lower errors across all test resolutions compared to the other competing methods, demonstrating its strong generalization capability and an excellent discretization-invariant property.

\section{Conclusion} 
\label{sec:conclusion}
In this work, we propose the Spectral Attention Operator Transformer (SAOT), a novel framework for operator learning designed at the attention mechanism from a complementary frequency-domain perspective. Considering the difficulties of Fourier attention in capturing high-frequency components with local details, we first introduce a novel Wavelet attention with linear complexity, which harnesses the spatial-frequency localization of wavelets to efficiently extract locality-aware features and compensate for the spectral limitation of Fourier attention. 
We further introduce a spectral attention layer that combines a Wavelet attention and a parallel Fourier attention via a gated fusion block to leverage the complementary advantages of these two spectral operations. 
Experimental results demonstrate that the proposed Wavelet attention offers valuable high-frequency information to recover the local details that Fourier attention inherently lacks. Hence, our SAOT model can achieve superior performance over other state-of-the-art methods. This work reveals the Wavelet transform's potential in operator learning, opening new research avenues such as exploring alternative wavelet basis variants.

\bibliography{myRefs}

@misc{tripura2022wavelet,
  title={Wavelet neural operator: a neural operator for parametric partial differential equations},
  author={Tripura, Tapas and Chakraborty, Souvik},
  eprint={2205.02191},
  archivePrefix={arXiv},
  year={2022}
}

@article{khan2018learning,
  title={Learning filter widths of spectral decompositions with wavelets},
  author={Khan, Haidar and Yener, Bulent},
  journal={Advances in Neural Information Processing Systems},
  volume={31},
  year={2018}
}

@inproceedings{NIPS2017_3f5ee243,
 author = {Vaswani, Ashish and Shazeer, Noam and Parmar, Niki and Uszkoreit, Jakob and Jones, Llion and Gomez, Aidan N and Kaiser, \L ukasz and Polosukhin, Illia},
 booktitle = {NeurIPS},
 pages = {},
 title = {Attention is All you Need},
 year = {2017}
}

@inproceedings{xiong2020layer,
  title={On layer normalization in the transformer architecture},
  author={Xiong, Ruibin and Yang, Yunchang and He, Di and Zheng, Kai and Zheng, Shuxin and Xing, Chen and Zhang, Huishuai and Lan, Yanyan and Wang, Liwei and Liu, Tieyan},
  booktitle={International conference on machine learning},
  pages={10524--10533},
  year={2020},
  organization={PMLR}
}

@inproceedings{hagnberger2024vectorized,
  title={Vectorized Conditional Neural Fields: A Framework for Solving Time-dependent Parametric Partial Differential Equations},
  author={Hagnberger, Jan and Kalimuthu, Marimuthu and Musekamp, Daniel and Niepert, Mathias},
  booktitle={ICML},
  year={2024}
}

@misc{wang2024bridging,
  title={Bridging Operator Learning and Conditioned Neural Fields: A Unifying Perspective},
  author={Wang, Sifan and Seidman, Jacob H and Sankaran, Shyam and Wang, Hanwen and Pappas, George J and Perdikaris, Paris},
  eprint={2405.13998},
  archivePrefix={arXiv},
  year={2024}
}

@inproceedings{kitaev2020reformer,
    title       = {Reformer: The Efficient Transformer},
    author      = {Nikita Kitaev and Lukasz Kaiser and Anselm Levskaya},
    booktitle   = {ICLR},
    year        = {2020},
}

@inproceedings{
choromanski2021rethinking,
title={Rethinking Attention with Performers},
author={Krzysztof Marcin Choromanski and Valerii Likhosherstov and David Dohan and Xingyou Song and Andreea Gane and Tamas Sarlos and Peter Hawkins and Jared Quincy Davis and Afroz Mohiuddin and Lukasz Kaiser and David Benjamin Belanger and Lucy J Colwell and Adrian Weller},
booktitle={ICLR},
year={2021},
}

@article{oommen2024integrating,
  author = {Oommen, Vivek  and Bora, Aniruddha  and Zhang, Zhen  and Karniadakis, George Em },
  title = {Integrating neural operators with diffusion models improves spectral representation in turbulence modelling},
  journal = {Proceedings of the Royal Society A: Mathematical, Physical and Engineering Sciences},
  volume = {481},
  number = {2309},
  pages = {20240819},
  year = {2025},
  doi = {10.1098/rspa.2024.0819}
}

@inproceedings{rahaman2019spectral,
  title={On the spectral bias of neural networks},
  author={Rahaman, Nasim and Baratin, Aristide and Arpit, Devansh and Draxler, Felix and Lin, Min and Hamprecht, Fred and Bengio, Yoshua and Courville, Aaron},
  booktitle={International conference on machine learning},
  pages={5301--5310},
  year={2019},
  organization={PMLR}
}

@inproceedings{cao2021towards,
  title={Towards Understanding the Spectral Bias of Deep Learning},
  author={Cao, Yuan and Fang, Zhiying and Wu, Yue and Zhou, Ding-Xuan and Gu, Quanquan},
  booktitle={30th International Joint Conference on Artificial Intelligence (IJCAI 2021)},
  pages={2205--2211},
  year={2021},
  organization={International Joint Conferences on Artificial Intelligence}
}

@misc{clevert2015fast,
  title={Fast and accurate deep network learning by exponential linear units ({ELUs})},
  author={Clevert, Djork-Arn{\'e}},
  eprint={1511.07289},
  archivePrefix={arXiv},
  year={2015}
}

@inproceedings{Katharopoulos2020TransformersAR,
  title={Transformers are RNNs: Fast Autoregressive Transformers with Linear Attention},
  author={Angelos Katharopoulos and Apoorv Vyas and Nikolaos Pappas and Franccois Fleuret},
  booktitle={ICML},
  year={2020}
}

@inproceedings{ronneberger2015u,
  title={U-net: Convolutional networks for biomedical image segmentation},
  author={Ronneberger, Olaf and Fischer, Philipp and Brox, Thomas},
  booktitle={International Conference on Medical image computing and computer-assisted intervention},
  pages={234--241},
  year={2015},
  organization={Springer}
}

@inproceedings{li2021fourier,
title={Fourier Neural Operator for Parametric Partial Differential Equations},
author={Zongyi Li and Nikola Borislavov Kovachki and Kamyar Azizzadenesheli and Burigede liu and Kaushik Bhattacharya and Andrew Stuart and Anima Anandkumar},
booktitle={ICLR},
year={2021}
}

@misc{lu2019deeponet,
  title={Deeponet: Learning nonlinear operators for identifying differential equations based on the universal approximation theorem of operators},
  author={Lu, Lu and Jin, Pengzhan and Karniadakis, George Em},
  eprint={1910.03193},
  archivePrefix={arXiv},
  year={2019}
}

@article{rahman2022u,
  title={U-no: U-shaped neural operators},
  author={Rahman, Md Ashiqur and Ross, Zachary E and Azizzadenesheli, Kamyar},
  journal={Transactions on Machine Learning Research},
  year={2023}
}

@inproceedings{wu2024Transolver,
  title={Transolver: A Fast Transformer Solver for {PDEs} on General Geometries},
  author={Haixu Wu and Huakun Luo and Haowen Wang and Jianmin Wang and Mingsheng Long},
  booktitle={ICML},
  year={2024}
}

@inproceedings{guibas2021adaptive,
  title={Adaptive fourier neural operators: Efficient token mixers for transformers},
  author={Guibas, John and Mardani, Morteza and Li, Zongyi and Tao, Andrew and Anandkumar, Anima and Catanzaro, Bryan},
  booktitle={ICLR},
  year={2022}
}

@article{jiang2023efficient,
  title={Efficient Super-Resolution of Near-Surface Climate Modeling Using the Fourier Neural Operator},
  author={Jiang, Peishi and Yang, Zhao and Wang, Jiali and Huang, Chenfu and Xue, Pengfei and Chakraborty, TC and Chen, Xingyuan and Qian, Yun},
  journal={Journal of Advances in Modeling Earth Systems},
  volume={15},
  number={7},
  year={2023},
  publisher={Wiley Online Library}
}

@misc{ba2016layer,
  title={Layer normalization},
  author={Ba, Jimmy Lei and Kiros, Jamie Ryan and Hinton, Geoffrey E},
  eprint={1607.06450},
  archivePrefix={arXiv},
  year={2016}
}

@misc{liu2024difffno,
  title={DiffFNO: Diffusion Fourier Neural Operator},
  author={Liu, Xiaoyi and Tang, Hao},
  eprint={2411.09911},
  archivePrefix={arXiv},
  year={2024}
}

@article{poli2022transform,
  title={Transform once: Efficient operator learning in frequency domain},
  author={Poli, Michael and Massaroli, Stefano and Berto, Federico and Park, Jinkyoo and Dao, Tri and R{\'e}, Christopher and Ermon, Stefano},
  journal={Advances in Neural Information Processing Systems},
  volume={35},
  pages={7947--7959},
  year={2022}
}

@inproceedings{hao2024dpot,
  title={{DPOT}: Auto-regressive denoising operator transformer for large-scale pde pre-training},
  author={Hao, Zhongkai and Su, Chang and Liu, Songming and Berner, Julius and Ying, Chengyang and Su, Hang and Anandkumar, Anima and Song, Jian and Zhu, Jun},
  booktitle={ICML},
  year={2024}
}

@book{bachman2000fourier,
  title={Fourier and wavelet analysis},
  author={Bachman, George and Narici, Lawrence and Beckenstein, Edward},
  volume={586},
  year={2000},
  publisher={Springer}
}

@inproceedings{chen2024positional,
               title={Positional Knowledge is All You Need: Position-induced Transformer ({PiT}) for Operator Learning},
               author={Junfeng Chen and Kailiang Wu},
               booktitle={ICML},
               year={2024}
}

@article{Wen2021UFNOA,
  title={U-FNO--An enhanced Fourier neural operator-based deep-learning model for multiphase flow},
  author={Wen, Gege and Li, Zongyi and Azizzadenesheli, Kamyar and Anandkumar, Anima and Benson, Sally M},
  journal={Advances in Water Resources},
  year={2022},
}

@misc{lee2021fnet,
  title={Fnet: Mixing tokens with fourier transforms},
  author={Lee-Thorp, James and Ainslie, Joshua and Eckstein, Ilya and Ontanon, Santiago},
  eprint={2105.03824},
  archivePrefix={arXiv},
  year={2021}
}

@inproceedings{rao2021global,
  title={Global filter networks for image classification},
  author={Rao, Yongming and Zhao, Wenliang and Zhu, Zheng and Lu, Jiwen and Zhou, Jie},
  booktitle={NeurIPS},
  year={2021}
}

@inproceedings{anonymous2023factorized,
title={Factorized Fourier Neural Operators},
author={Alasdair Tran and Alexander Mathews and Lexing Xie and Cheng Soon Ong},
booktitle={ICLR},
year={2023}
}

@inproceedings{Gupta2021MultiwaveletbasedOL,
  title={Multiwavelet-based Operator Learning for Differential Equations},
  author={Gaurav Gupta and Xiongye Xiao and Paul Bogdan},
  booktitle={NeurIPS},
  year={2021}
}

@article{lu2021learning,
  title={Learning nonlinear operators via DeepONet based on the universal approximation theorem of operators},
  author={Lu, Lu and Jin, Pengzhan and Pang, Guofei and Zhang, Zhongqiang and Karniadakis, George Em},
  journal={Nature machine intelligence},
  volume={3},
  number={3},
  pages={218--229},
  year={2021},
}

@article{Li2022FourierNO,
  title={Fourier Neural Operator with Learned Deformations for {PDEs} on General Geometries},
  author={Zong-Yi Li and Daniel Z. Huang and Burigede Liu and Anima Anandkumar},
  journal={Journal of Machine Learning Research},
  volume={24},
  number={388},
  pages={1--26},
  year={2023}
}

@article{kovachki2023neural,
  title={Neural operator: Learning maps between function spaces with applications to {PDEs}},
  author={Kovachki, Nikola and Li, Zongyi and Liu, Burigede and Azizzadenesheli, Kamyar and Bhattacharya, Kaushik and Stuart, Andrew and Anandkumar, Anima},
  journal={Journal of Machine Learning Research},
  volume={24},
  number={89},
  pages={1--97},
  year={2023}
}

@inproceedings{Cao2021ChooseAT,
  title={Choose a Transformer: Fourier or Galerkin},
  author={Shuhao Cao},
  booktitle={NeurIPS},
  year={2021}
}

@misc{Fanaskov2022SpectralNO,
  title={Spectral Neural Operators},
  author={Vladimir Fanaskov and I. Oseledets},
  eprint={2205.10573},
  archivePrefix={arXiv},
  year={2022}
}

@book{roubivcek2013nonlinear,
  title={Nonlinear partial differential equations with applications},
  author={Roub{\'\i}{\v{c}}ek, Tom{\'a}{\v{s}}},
  year={2013},
  publisher={Springer Science \& Business Media}
}

@misc{Li2020NeuralOG,
  title={Neural Operator: Graph Kernel Network for Partial Differential Equations},
  author={Zong-Yi Li and Nikola B. Kovachki and Kamyar Azizzadenesheli and Burigede Liu and Kaushik Bhattacharya and Andrew Stuart and Anima Anandkumar},
  eprint={2003.03485},
  archivePrefix={arXiv},
  year={2020}
}

@inproceedings{wu2023LSM,
  title={Solving High-Dimensional PDEs with Latent Spectral Models},
  author={Haixu Wu and Tengge Hu and Huakun Luo and Jianmin Wang and Mingsheng Long},
  booktitle={ICML},
  year={2023}
}

@article{azizzadenesheli2024neural,
  title={Neural operators for accelerating scientific simulations and design},
  author={Azizzadenesheli, Kamyar and Kovachki, Nikola and Li, Zongyi and Liu-Schiaffini, Miguel and Kossaifi, Jean and Anandkumar, Anima},
  journal={Nature Reviews Physics},
  pages={1--9},
  year={2024},
  publisher={Nature Publishing Group UK London}
}

@inproceedings{yao2022wave,
  title={Wave-vit: Unifying wavelet and transformers for visual representation learning},
  author={Yao, Ting and Pan, Yingwei and Li, Yehao and Ngo, Chong-Wah and Mei, Tao},
  booktitle={ECCV},
  pages={328--345},
  year={2022},
  organization={Springer}
}

@inproceedings{hao2023gnot,
  title={{GNOT}: A General Neural Operator Transformer for Operator Learning},
  author={Hao, Zhongkai and Ying, Chengyang and Wang, Zhengyi and Su, Hang and Dong, Yinpeng and Liu, Songming and Cheng, Ze and Zhu, Jun and Song, Jian},
  booktitle={ICML},
  year={2023}
}

@article{
li2023transformer,
title={Transformer for Partial Differential Equations{\textquoteright} Operator Learning},
author={Zijie Li and Kazem Meidani and Amir Barati Farimani},
journal={Transactions on Machine Learning Research},
year={2023},
}

@inproceedings{li2023scalable,
      title={Scalable Transformer for PDE Surrogate Modeling}, 
      booktitle={NeurIPS},
      author={Zijie Li and Dule Shu and Amir Barati Farimani},
      year={2023},
}

@inproceedings{
anonymous2023improved,
title={Improved Operator Learning by Orthogonal Attention},
author={Zipeng Xiao and Zhongkai Hao and Bokai Lin and Zhijie Deng and Hang Su},
booktitle={ICML},
year={2024},
}

@inproceedings{liu2020wavelet,
  title={Wavelet-based dual-branch network for image demoir{\'e}ing},
  author={Liu, Lin and Liu, Jianzhuang and Yuan, Shanxin and Slabaugh, Gregory and Leonardis, Ale{\v{s}} and Zhou, Wengang and Tian, Qi},
  booktitle={ECCV},
  year={2020},
}

@misc{yue2024point,
  title={Point-Calibrated Spectral Neural Operators},
  author={Yue, Xihang and Zhu, Linchao and Yang, Yi},
  eprint={2410.11382},
  archivePrefix={arXiv},
  year={2024}
}

@inproceedings{liu2024neural,
  title={Neural operators with localized integral and differential kernels},
  author={Liu-Schiaffini, Miguel and Berner, Julius and Bonev, Boris and Kurth, Thorsten and Azizzadenesheli, Kamyar and Anandkumar, Anima},
  booktitle={ICML},
  year={2024}
}

@inproceedings{katharopoulos2020transformers,
  title={Transformers are rnns: Fast autoregressive transformers with linear attention},
  author={Katharopoulos, Angelos and Vyas, Apoorv and Pappas, Nikolaos and Fleuret, Fran{\c{c}}ois},
  booktitle={ICML},
  year={2020},
}

@article{li2024component,
  title={Component fourier neural operator for singularly perturbed differential equations},
  author={Li, Ye and Du, Ting and Pang, Yiwen and Huang, Zhongyi},
  journal={Proceedings of the AAAI Conference on Artificial Intelligence},
  volume={38},
  number={12},
  pages={13691--13699},
  year={2024}
}

@article{cho2024operator,
  title={Operator-learning-inspired modeling of neural ordinary differential equations},
  author={Cho, Woojin and Cho, Seunghyeon and Jin, Hyundong and Jeon, Jinsung and Lee, Kookjin and Hong, Sanghyun and Lee, Dongeun and Choi, Jonghyun and Park, Noseong},
  journal={Proceedings of the AAAI Conference on Artificial Intelligence},
  volume={38},
  number={10},
  pages={11543--11551},
  year={2024}
}

@article{lee2024inducing,
  title={Inducing point operator transformer: A flexible and scalable architecture for solving pdes},
  author={Lee, Seungjun and Oh, Taeil},
  journal={Proceedings of the AAAI Conference on Artificial Intelligence},
  volume={38},
  number={1},
  pages={153--161},
  year={2024}
}

@article{leng2025fnin,
  title={FNIN: A Fourier Neural Operator-based Numerical Integration Network for Surface-from-gradients},
  author={Leng, Jiaqi and Ju, Yakun and Duan, Yuanxu and Zhang, Jiangnan and Lv, Qingxuan and Wu, Zuxuan and Fan, Hao},
  journal={Proceedings of the AAAI Conference on Artificial Intelligence},
  volume={39},
  number={5},
  pages={4580--4588},
  year={2025}
}

@article{liu2025aerogto,
  title={Aerogto: An efficient graph-transformer operator for learning large-scale aerodynamics of 3d vehicle geometries},
  author={Liu, Pengwei and Wang, Pengkai and Ren, Xingyu and Yuan, Hangjie and Hao, Zhongkai and Xu, Chao and Cai, Shengze and Ni, Dong},
  journal={Proceedings of the AAAI Conference on Artificial Intelligence},
  volume={39},
  number={18},
  pages={18924--18932},
  year={2025}
}

\end{document}